%% file: main.tex
\documentclass[10pt,twocolumn,letterpaper]{article}

\usepackage[pagenumbers]{cvpr} 


\definecolor{cvprblue}{rgb}{0.21,0.49,0.74}
\usepackage[pagebackref,breaklinks,colorlinks,allcolors=cvprblue]{hyperref}
\usepackage{bm}
\usepackage{algorithm}
\usepackage{algorithmic}
\usepackage{pifont}
\usepackage[accsupp]{axessibility}
\def\paperID{3755} 
\def\confName{CVPR}
\def\confYear{2025}

\title{Instant Adversarial Purification with Adversarial Consistency Distillation}

\author{Chun Tong Lei$^1$, Hon Ming Yam$^1$, Zhongliang Guo$^2$, Yifei Qian$^3$, Chun Pong Lau$^1$\thanks{Corresponding Author.}\\
{\normalsize$^1$City University of Hong Kong} \quad {\normalsize$^2$University of St Andrews} \quad {\normalsize$^3$University of Nottingham}\\
{\small\{ctlei2, hmyam4, cplau27\}@cityu.edu.hk,}
{\small zg34@st-andrews.ac.uk, yifei.qian@nottingham.ac.uk
    }}

\begin{document}
\maketitle

\input{sec/0_abstract}

\setlength{\abovedisplayskip}{3pt}
\setlength{\belowdisplayskip}{3pt}
\setlength{\abovecaptionskip}{3pt}
\setlength{\belowcaptionskip}{1pt}
\setlength{\textfloatsep}{1pt}
\input{sec/1_intro}
\input{sec/2_related}
\input{sec/3_prelim}
\input{sec/4_method}

\input{sec/5_exp}
\input{sec/6_conclu}
\clearpage
{
    \small
    \bibliographystyle{ieeenat_fullname}
    \bibliography{main}
}

\input{sec/X_suppl}

\end{document}

%% file: sec/0_abstract.tex
\begin{abstract}
Neural networks have revolutionized numerous fields with their exceptional performance, yet they remain susceptible to adversarial attacks through subtle perturbations. While diffusion-based purification methods like DiffPure offer promising defense mechanisms, their computational overhead presents a significant practical limitation.
In this paper, we introduce One Step Control Purification (OSCP), a novel defense framework that achieves robust adversarial purification in a single Neural Function Evaluation (NFE) within diffusion models.
We propose Gaussian Adversarial Noise Distillation (GAND) as the distillation objective and Controlled Adversarial Purification (CAP) as the inference pipeline, which makes OSCP demonstrate remarkable efficiency while maintaining defense efficacy.
Our proposed GAND addresses a fundamental tension between consistency distillation and adversarial perturbation, bridging the gap between natural and adversarial manifolds in the latent space, while remaining computationally efficient through Parameter-Efficient Fine-Tuning (PEFT) methods such as LoRA, eliminating the high computational budget request from full parameter fine-tuning.
The CAP guides the purification process through the unlearnable edge detection operator calculated by the input image as an extra prompt, effectively preventing the purified images from deviating from their original appearance when large purification steps are used.
Our experimental results on ImageNet showcase OSCP's superior performance, achieving a 74.19\% defense success rate with merely 0.1s per purification --- a 100-fold speedup compared to conventional approaches.
\end{abstract}

%% file: sec/1_intro.tex
\section{Introduction}
\label{sec:intro}

\begin{figure}[htbp]
    \centering
    \includegraphics[width=\linewidth]{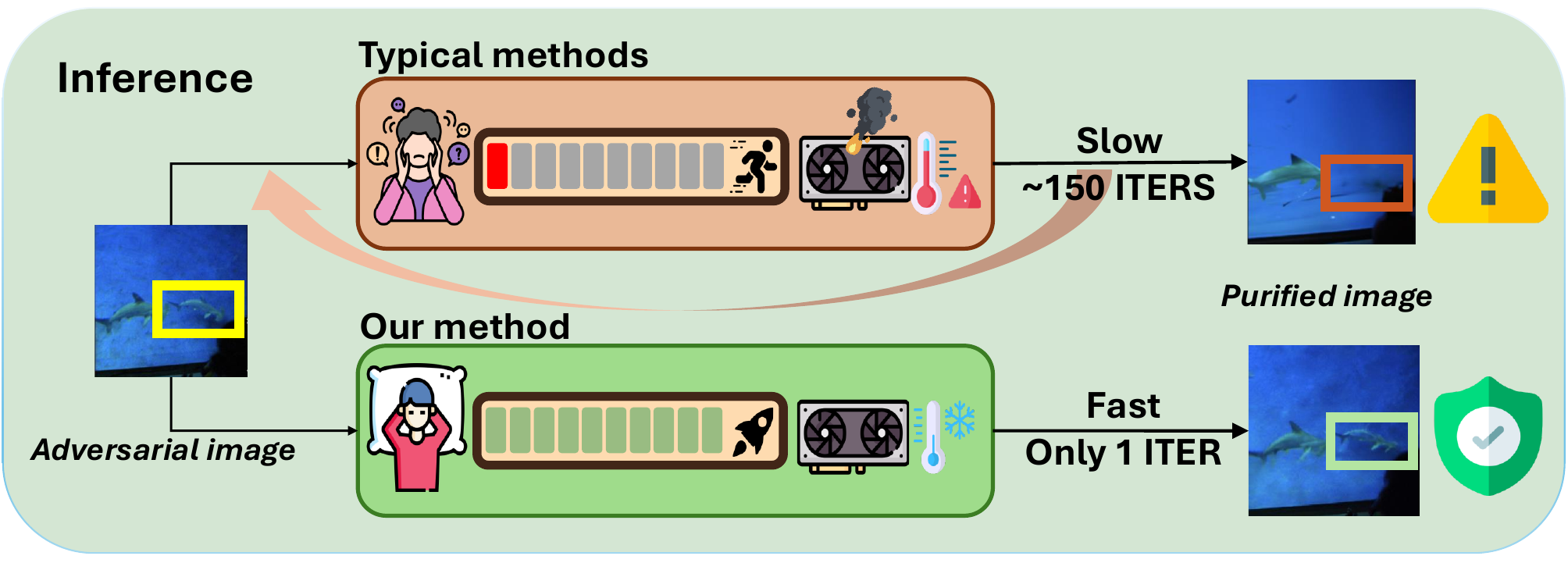}\\
    \vspace{0.7em}
    \includegraphics[width=0.32\linewidth]{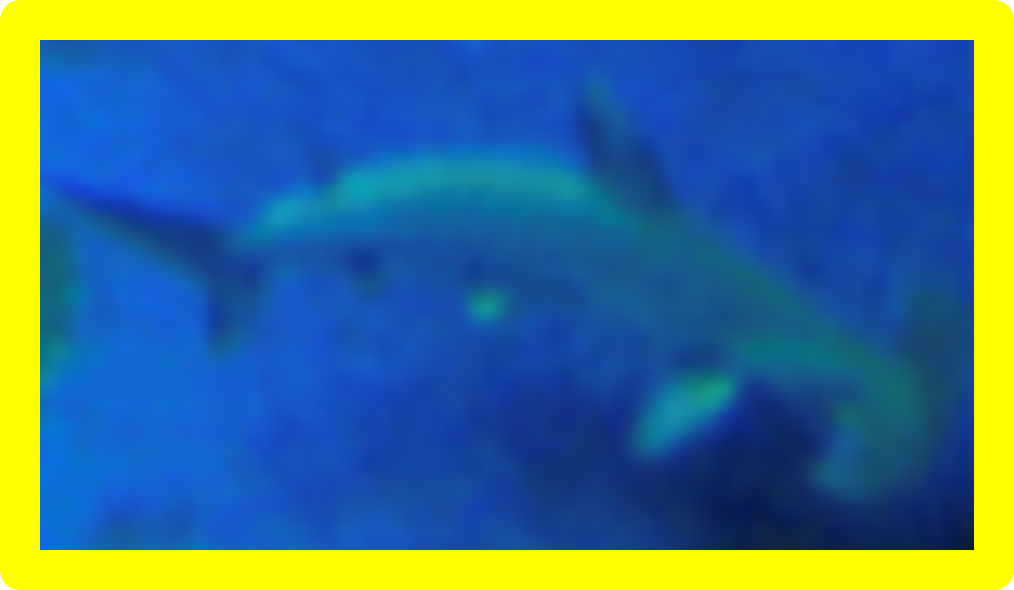}
    \includegraphics[width=0.32\linewidth]{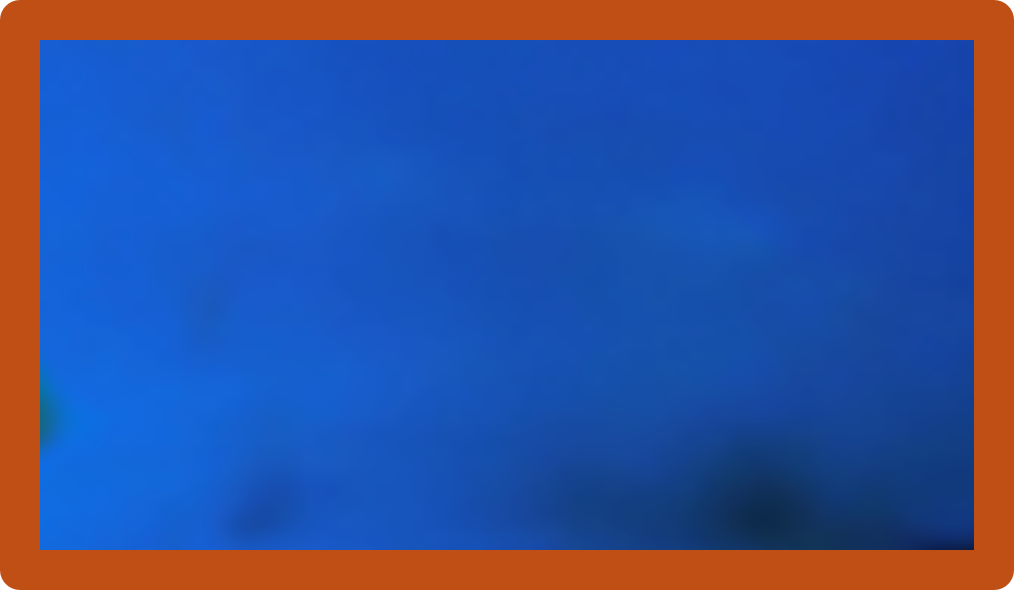}
    \includegraphics[width=0.32\linewidth]{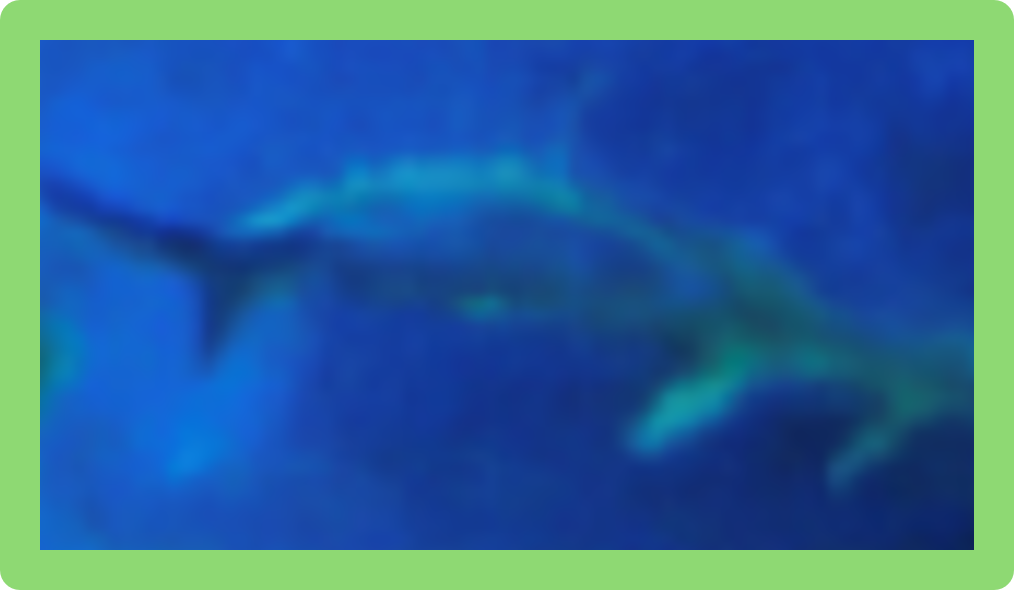}
    \caption{Comparison between existing methods and our proposed approach. Our method achieves superior performance with just a single inference step, significantly reducing computational cost. Through adversarial noise-adapted U-Net fine-tuning, we demonstrate better detail preservation after denoising, as evident in the zoomed-in regions for circles (bottom). This makes OSCP an efficient and practical solution for adversarial purification.}
    \label{fig:teaser}
    \vspace{1mm}
\end{figure}

\begin{figure*}[ht]
    \centering
    \includegraphics[width=0.4\textwidth]{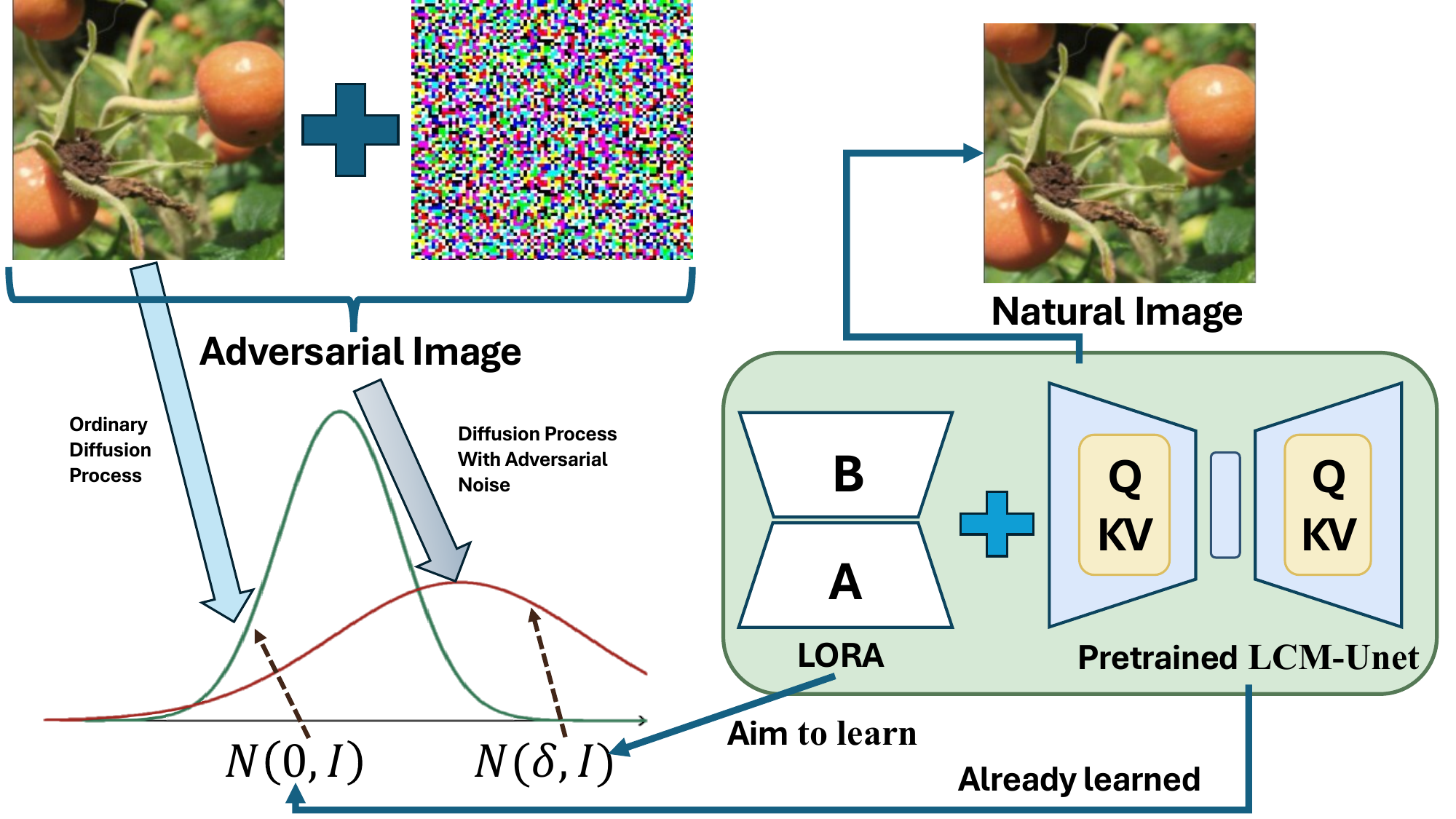}
    \hspace{0.001\textwidth}\vrule\hspace{0.005\textwidth}
    \includegraphics[width=0.58\textwidth]{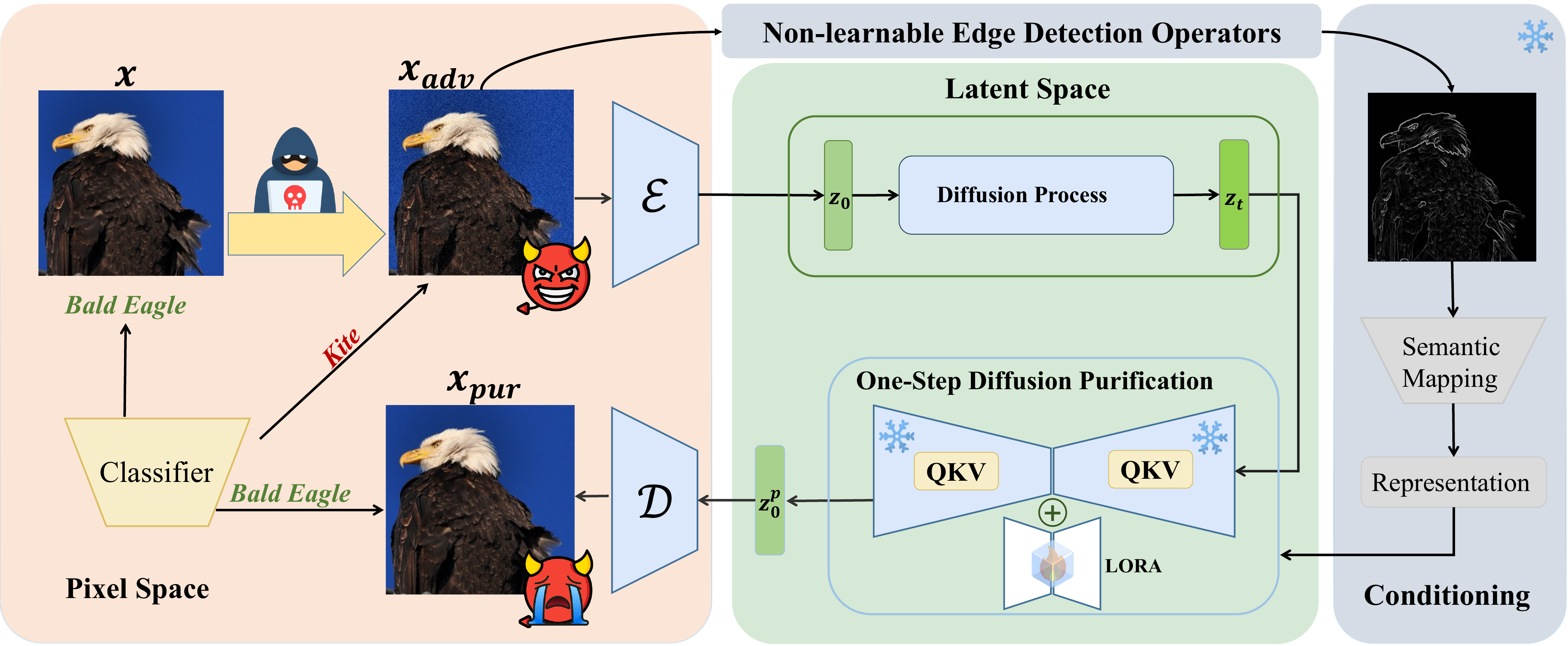}
    \caption{The pipeline of Our proposed OSCP.
    \textbf{(a)} the left figure shows that the adversarial images, which are crafted through intentional attacks, exhibit a shifted distribution after the diffusion process that deviates from the Standard Normal distribution. In response, our proposed GAND can learn to recover the attacked images by modeling this additional adversarial noise with LoRA.
    \textbf{(b)} the right figure illustrates the pipeline that our proposed CAP leverages non-learnable edge detection operators to guide the purification of adversarial samples, avoiding potential inductive bias introduced by neural networks.
    It is worth noting that our method achieves remarkable performance by just running a single U-Net inference step.}
    \label{fig:pipe}
\end{figure*}

Deep Neural Networks (DNNs) have fundamentally transformed the landscape of computer vision, achieving unprecedented performance across diverse applications. Despite these extraordinary achievements, a critical vulnerability lurks beneath their impressive performance, one that has emerged as a paramount concern in the research community~\cite{madry2018towards,guo2024grey,sam-tifs-adv-training}.

The Achilles' heel of these sophisticated neural architectures lies in their remarkable vulnerability to adversarial attacks~\cite{szegedy2013intriguing,sam-instruct2attack}. Through the injection of meticulously engineered perturbations that are imperceptible to human observers yet devastating to model performance, adversaries can systematically manipulate DNNs into catastrophic misclassifications~\cite{guo2024white}.
While this vulnerability has found constructive applications in safeguarding against AI misuse, such as protecting intellectual property and preventing unauthorized model exploitation~\cite{guo2024artwork}, its implications extend far beyond these beneficial uses. The susceptibility to adversarial manipulation poses a critical threat to the deployment of DNNs in high-stakes domains, where system reliability and robustness are not merely desirable but essential for public safety and security~\cite{sam-indentify-adv}.

In response to these security challenges, two principal defense paradigms have emerged: adversarial training~\cite{wangbetter,sam-nips-defense} and adversarial purification~\cite{samangouei2018defense,yoon2021adversarial,nie2022diffusion}. Adversarial training, despite its widespread adoption, exhibits a critical limitation --- it requires prior knowledge of attack methods, inherently constraining its effectiveness against unknown threats. In contrast, adversarial purification offers a more versatile defense framework by focusing on perturbation removal rather than attack-specific training, making it inherently more adaptable to emerging adversarial threats.

Adversarial purification methods, particularly those leveraging diffusion models~\cite{ho2020denoising,songdenoising}, have demonstrated remarkable effectiveness in neutralizing adversarial attacks by mapping perturbed samples back to the natural distribution~\cite{nie2022diffusion,wang2022guided}. Diffusion models, with their unique denoising capabilities and superior training stability, offer significant advantages over traditional generative approaches such as GANs. Their ability to systematically restore corrupted images through iterative denoising has established them as a compelling framework for adversarial defense.

Despite their effectiveness, diffusion models face a critical limitation:\textbf{ their multi-step denoising process incurs substantial computational overhead}, making them impractical for real-time defense scenarios. This inherent inefficiency poses a significant challenge in applications demanding rapid response times, such as real-time security systems. Consequently, there is an urgent need for efficient purification methods that can maintain robust defense while minimizing computational costs.

To address these challenges, we introduce One Step Control Purification (OSCP), a novel diffusion-based framework that achieves robust adversarial defense in a single inference step. At its core, OSCP first leverages consistency distillation for fast inference while explicitly addressing the fundamental discrepancy between adversarial and clean sample distributions. This insight leads to our first key innovation, the distillation objective --- Gaussian Adversarial Noise Distillation (GAND). As shown in Fig.~\ref{fig:teaser} and Fig.~\ref{fig:pipe} (left), GAND enables efficient purification while maintaining strong defense performance.

Our work also addresses another fundamental challenge in diffusion-based purification --- \textbf{the loss of semantic information in large-step inference}. While Latent Consistency Models offer acceleration, they often sacrifice image quality, resulting in semantic degradation and blurry outputs. To preserve both efficiency and fidelity, as shown in Fig.~\ref{fig:pipe} (right), we propose Controlled Adversarial Purification (CAP) as the inference pipeline to empower GAND, which integrates ControlNet with non-learnable edge detection operators, enabling high-quality purification while maintaining semantic integrity.

Our proposed OSCP framework, composed of GAND as the distillation objective and CAP as the inference pipeline, achieves state-of-the-art performance with 74.19\% robust accuracy on ImageNet while requiring merely 0.1s per purification. This breakthrough in efficiency and effectiveness not only advances the field of adversarial defense but also enables the practical deployment of robust neural networks in real-world, time-sensitive applications.

In summary, our key contributions are as follows:
\begin{itemize}
    \item We propose GAND, a novel consistency distillation objective for adversarial training on Latent Consistency Models. Our empirical results demonstrate its exceptional robustness and transferability against unknown adversarial attacks.
    \item We introduce CAP, an innovative inference pipeline that leverages non-learnable edge detection operators to enhance the controllability and semantic preservation of adversarial purification.
    \item We develop OSCP, an integrated framework combining GAND and CAP, that achieves real-time adversarial purification while maintaining robust defense performance, making diffusion-based defenses practical for time-critical applications.
\end{itemize}

%% file: sec/2_related.tex
\section{Related Work}
\label{sec:relatedwork}
\subsection{Adversarial training}
Adversarial training~\cite{zhao2025a} has established itself as a cornerstone defense strategy against adversarial attacks by incorporating perturbed examples into the training process~\cite{sam-pami-defense}. Numerous studies have shown its effectiveness~\cite{athalye2018obfuscated,gowal2020uncovering}, showing significant improvements in model robustness against known attack patterns. However, this approach exhibits an inherent limitation: models tend to overfit known attacks, compromising their resilience against novel attack vectors~\cite{sam-nips-defense}. Recent advances attempt to address this limitation by leveraging diffusion models to generate diverse adversarial samples~\cite{gowal2021improving,wangbetter}, aiming to enhance generalization and prevent overfitting. 

\subsection{Adversarial Purification} Adversarial purification has evolved significantly in its approach to defending against attacks through input restoration. Initially pioneered with GAN-based methods~\cite{samangouei2018defense}, the field underwent a paradigm shift with the advent of diffusion models~\cite{song2019generative,ho2020denoising,songdenoising}. These diffusion-based approaches have demonstrated superior purification capabilities~\cite{yoon2021adversarial,nie2022diffusion,wang2022guided}. Notably, DiffPure~\cite{nie2022diffusion} simultaneously removes adversarial perturbations and Gaussian noise from the forward process, theoretically justified by the reduced KL divergence between clean and adversarial image distributions. Despite these advances, the computational overhead of multiple denoising steps remains a critical bottleneck~\cite{wang2022guided}, limiting practical deployment.

\subsection{Diffusion Models} Diffusion models have demonstrated remarkable capabilities across diverse domains, including text-to-image synthesis~\cite{rombach2022high,saharia2022photorealistic}, video generation~\cite{ho2022video,blattmann2023align}, and 3D content creation~\cite{luo2021diffusion,poole2022dreamfusion}. The foundational Denoising Diffusion Probabilistic Models (DDPM)~\cite{ho2020denoising} established a two-phase framework: a forward process that gradually applies Gaussian noise following Markov properties, and a reverse process that learns to reconstruct the original image through a reverse Markov chain. However, the sequential nature of this process results in prohibitively long inference times, as computational cost scales linearly with the number of denoising steps.

\subsection{Efficient Diffusion Models} To address the computational bottleneck, Consistency Models~\cite{song2023consistency} introduced a novel consistency training paradigm, enabling image generation in significantly fewer steps. This advancement led to the development of Latent Consistency Models (LCM)~\cite{luo2023latent}, which further accelerated the generation process. LCM-LoRA~\cite{luo2023lcm} enhanced efficiency through Low-Rank Adaptation, a Parameter-Efficient Fine-Tuning (PEFT)~\cite{houlsby2019parameter} approach that substantially reduces computational requirements. Additionally, ControlNet~\cite{zhang2023adding} introduced flexible conditioning mechanisms for Stable Diffusion, enabling precise control over generation through various visual guides including edge maps, pose estimation, sketches, and depth information.

%% file: sec/3_prelim.tex
\section{Preliminaries}
\label{sec:preliminaries}

\subsection{Diffusion model}
Denoising Diffusion Probabilistic Models (DDPM)~\cite{ho2020denoising} generate images by learning from the reverse Markov chain with Gaussian noise added to the original image. The forward process can be formulated as linear combination of original image $\bm{x}_0$ and standard Gaussian noise $\bm{\epsilon}$, $\Bar{\alpha}_t$ denoted cumulative product from $\alpha_1$ to $\alpha_t$, $\alpha_t=1-\beta_t$ for any $t$, $\beta_t$ is predefined variance schedule of diffusion process:
\begin{equation}
    \bm{x}_t = \sqrt{\Bar{\alpha}_t}\bm{x}_0+\sqrt{1-\Bar{\alpha}_t}\bm{\epsilon}, \quad \bm{\epsilon}\sim \mathcal{N}(0,\mathbf{I}),
    \label{eq:ddpm}
\end{equation}
and the denoising step can be expressed as:
\begin{equation}
    \hat{\bm{x}}_{t-1}=\frac{1}{\sqrt{\alpha_t}}\left({\bm{x}_t-\frac{\beta_t}{\sqrt{1-\Bar{\alpha}_t}}}\bm{\epsilon}_{\bm{\theta^*}}(\bm{x}_t,t)\right)+\sqrt{\beta_t}\bm{\epsilon},
\end{equation}
where model parameter $\bm{\theta}^\ast$ minimize the loss between actual noise and predict noise:
\begin{equation}
    \bm{\theta}^\ast=\arg\min_\theta\mathbb{E}_{\bm{x}_0,t,\bm{\epsilon}} \left[\lVert\bm{\epsilon}-\bm{\epsilon}_\theta(\bm{x}_t,t)\lVert_2^2\right].
\end{equation}

\subsection{Diffusion-Base Purification (DBP)}
DiffPure~\cite{nie2022diffusion} proposes that diffusion models can remove adversarial noise by performing a sub-process of the normal reverse process ($t=0$ to $t=T$). By adding predefined $t^\ast$ ($t^\ast<T$) of noise to the adversarial image $\bm{x}_{adv}$, which is formed by the sum of original image $\bm{x}$ and adversarial noise $\bm{\delta}$, $\bm{\delta}$ can be generated by $L_{p}$ attack~\cite{madry2018towards} or AutoAttack~\cite{croce2020reliable}:
\begin{equation}
    \bm{\delta}=\arg\max_\delta\mathcal{L}(C(\bm{x}+\bm{\delta}),y),
\end{equation}
where $C$ is the classifier, $y$ is the true label. The forward process of diffusion purification method uses:
\begin{equation}
    \bm{x}(t^\ast) = \sqrt{\Bar{\alpha}(t^\ast)}\bm{x}+\sqrt{1-\Bar{\alpha}(t^\ast)}\bm{\epsilon},
\end{equation}
and solve the reverse process of DDPM from time step $t^\ast$ to $0$, to get the purified $\hat{\bm{x}}_{adv}^0$ that is close to the original image $\bm{x}$, allowing the classifier to classify the image with the correct label.

\subsection{Consistency Model}

Diffusion models are known to have long inference time, which limits their usage in the real world. To tackle this situation, Consistency model~\cite{song2023consistency} has been proposed, which makes it possible to generate images within 2 to 4 U-Net inference steps, by distilling a Consistency Model from a pretrianed diffusion model.

Due to the efficiency of latent space models compared to pixel-based models, Latent Consistency Model~\cite{luo2023latent} has been proposed, making use of pretrained encoder and decoder to transform images from pixel space to latent space. Latent Consistency Model's definition is 
\begin{equation}
   f_\theta(\bm{z}_t,t)=f_\theta(\bm{z}_{t'},t')\quad\forall t,t' \in [0,T]
   \label{consistency function defintion}
\end{equation} and $f_\theta$ can be parameterized as:
\begin{equation}
   f_{\theta}(\bm{z},\bm{c},t)=c_\text{skip}(t)\bm{z}+c_\text{out}(t)\left(\frac{\bm{z}-\sqrt{1-\Bar{\alpha}_t}\hat{\bm\epsilon}_\theta(\bm{z},\bm{c},t)}{\sqrt{\Bar{\alpha}_t}}\right),
   \label{latent consistency function}
\end{equation}
where $c_\text{skip}(t)$ and $c_\text{out}(t)$ are differentiable functions such that $c_\text{skip}(\epsilon)=1$ and $c_\text{out}(\epsilon)=0$. Distilled with LoRA is introduced in LCM-LoRA~\cite{luo2023lcm}, which dramatically reduces training time and computation cost in distillation by hugely reducing trainable parameters. 

%% file: sec/4_method.tex
\section{Method}

\newtheorem{theorem}{Theorem}

In this paper, we aim to solve the slow purification problem in DBP by leveraging the Latent Consistency Distillation to speed up the purification backbone, enabling single-step adversarial purification.

Recognizing that adversarial noise differs from the Gaussian noise that diffusion models are typically designed to remove, we propose Gaussian Adversarial Noise Distillation (GAND). This novel distillation objective specifically targets adversarial noise, enhancing the purification performance. Our approach builds upon the insight that combining adversarial purification with adversarial training can yield superior results~\cite{liu2024towards}, effectively addressing the distinct distributions of Gaussian and adversarial noise.

However, diffusion models tend to produce images deviating from the originals when purification steps $t^\ast$ are large~\cite{wang2022guided}. To address this, we introduce Controlled Adversarial Purification (CAP), shown in Fig.~\ref{fig:pipe} (right), which utilizes the unlearnable edge detection operator computed on the input to guide the purification process.

\subsection{One Step Control Purification}

\paragraph{Problem Definition}
Our goal can be formulated as:

\begin{equation}
\begin{aligned}
    \bm{x}_{gt}\simeq\hat{\bm{x}}_{adv}^0=\mathcal{D}(f(\mathcal{E}(\bm{x}_{adv}),t^\ast))&\\
    \text{s.t.}\quad C(\bm{x}_{gt})=C(\hat{\bm{x}}_{adv}^0)&,
\end{aligned}
\end{equation}
where $\bm{x}_{gt}$ is groundtruth image, $\bm{x}_{adv}$ is the adversarial image, $f$ is a model with the purification function in latent space, $t^\ast$ is predefined purification step, $\mathcal{E}$ and $\mathcal{D}$ are pretrained image encoder and decoder respectively. For simplifying symbolic, we denote $\mathcal{E}(\bm{x}_{adv})$ as $\bm{z}_{adv}$.

An overview of our method OSCP is illustrated in Fig.~\ref{fig:pipe}; OSCP can be separated into two components: a) training a sped-up backbone model with noise and denoise function; b) pipeline for purification process with non-text guidance. We propose Gaussian Adversarial Noise Distillation (GAND) and Controlled Adversarial Purification (CAP) to respond to those two unique requirements.

\subsection{Gaussian Adversarial Noise Distillation}
To respond to the first component we defined, we propose GAND as illustrated in Fig.~\ref{fig:pipe} (left) and detailed in Alg.~\ref{alg:adv_lcm}, a novel adversarial distillation objective, which combines adversarial training, Latent Consistency Distillation, and learning how to remove the adversarial and Gaussian noise from the shifted standard Gaussian distribution.

\begin{algorithm}[tb]
\caption{GAND}\label{alg:adv_lcm}
\begin{algorithmic}[1]
\REQUIRE dataset $\mathcal{X}$, class label of the image $\bm{y}$, classifier $C$, LCM $f_{\theta}$ and it’s parameter $\bm{\theta}$, ODE solver $\Psi$ and distance metric $d$. $\mathcal{L}_{C}$ and $\mathcal{L}_{G}$ indicate $\mathcal{L}_{CIG}$ (\cref{cig}) and $\mathcal{L}_{GAND}$ (\cref{gand}) respectively. $t'$ denote time step $ t_{n+k}$ in diffusion process.
\WHILE{not convergence}
\STATE Sample $\bm{x}\sim\mathcal{X}$, $n\sim\mathcal{U}[1,(N-k)/2]$
\STATE $\bm{z}=\mathcal{E}(\bm{x})$
\STATE $\bm{\delta}_{adv}=\arg\max_\delta L(C(\mathcal{D}(\bm{z}+\bm{\delta})),\bm{y})$ \hfill~\cref{latent space attack}
\STATE $\bm{z}^*_{t'}=\sqrt{\bar{\alpha}_{t'}}\bm{z}+\sqrt{1-\bar{\alpha}_{t'}}(\bm{\epsilon}+\bm{\delta}_{adv})$ \hfill~\cref{forward diffuse}
\STATE $\hat{\bm{z}}_{t_n}^{\Psi} \gets \bm{z}^*_{t'}+\Psi(\bm{z}^*_{t'},t',t_n,\varnothing)$ \hfill~\cref{denoise step}
\STATE $\mathcal{L}_{G}(\bm{\theta},\bm{\theta}^-)\gets d(f_\theta(\bm{z}^*_{t'},\varnothing,t'),f_{\theta^-}(\hat{\bm{z}}_{t_n}^{\Psi},\varnothing,t_n))$
\STATE $\mathcal{L}_{C}(\bm{\theta})\gets d(f_\theta(\bm{z}^*_{t'},\varnothing,t'),\bm{z})$
\STATE $\mathcal{L}_{T}(\bm{\theta},\bm{\theta}^-)\gets \mathcal{L}_{G}(\bm{\theta},\bm{\theta}^-)+\lambda_{CIG}\mathcal{L}_{C}(\bm{\theta})$  \hfill~\cref{total loss}
\STATE $\bm{\theta}\gets\bm{\theta}-\eta\nabla_\theta\mathcal{L}(\bm{\theta},\bm{\theta}^-)$
\ENDWHILE

\end{algorithmic}

\end{algorithm}

The original latent consistency distillation first encodes the image to latent code by encoder $\mathcal{E}$, such that $\bm{z}=\mathcal{E}(\bm{x})$ and trains the model base on loss function:
\begin{equation}
\begin{split}
    \mathcal{L}(\theta,&\theta^-;\Psi)=\\ &\mathbb{E}_{\bm{z},\bm{c},n}[d(f_\theta(\bm{z}_{t_{n+k}}, \bm{c},t_{n+k}),f_{\theta^-}(\hat{\bm{z}}^{\Psi,\omega}_{t_{n}},\bm{c},t_{n}))]
    \label{distilation loss},
\end{split}
\end{equation}
where $d(\cdot,\cdot)$ is distance metric, $\Psi(\cdot, \cdot, \cdot, \cdot) $ is DDIM PF-ODE solver $\Psi_\text{DDIM}$~\cite{luo2023latent}
, $f(\cdot,\cdot,\cdot)$ is latent consistency function in E.q.~\ref{latent consistency function} and $\hat{\bm{z}}^{\Psi,\omega}_{t_{n}}$ is an estimation of the solution of the PF-ODE from $t_{n+k} \rightarrow t_n$ using the DDIM PF-ODE solver $\Psi$:
\begin{equation}
\begin{split}
        \hat{\bm{z}}^{\Psi,\omega}_{t_{n}}\leftarrow\bm{z}_{t_{n+k}}+(1+\omega)\Psi&(\bm{z}_{t_{n+k}},t_{n+k},t_n,c)\\ -\omega&\Psi(\bm{z}_{t_{n+k}},t_{n+k},t_n,\varnothing),
\end{split}
\end{equation}
$\omega$ is guidance scale and it is sampled from $[\omega_{\min} , \omega_{\max}]$.

We aim to learn the denoise trajectory from a combination of Gaussian noise and adversarial noise to natural image; we need a model satisfying that $f_\theta(\bm{z}_{adv}(t),\varnothing,t)=f_\theta(\bm{z}(t),\varnothing,t)=\bm{z}_\epsilon$ for all $t$. $\bm{z}_{\epsilon}$ is the limit of $\bm{z}_t$ when $t\rightarrow0$ where $\bm{z}_{\epsilon}\simeq \bm{z}$. However, if we look closely at the formulation of $f_\theta$ in ~\cref{latent consistency function}, $f_\theta(\bm{z}_{adv}(t),\varnothing,t)$ is converging to $\bm{z}_{adv}$ when $t\rightarrow 0$. $f_\theta(\bm{z}_{adv}(t),\varnothing,t)-f_\theta(\bm{z}(t),\varnothing,t)\rightarrow\bm{\delta}_{adv}$ when $t\rightarrow 0$, where $\bm{\delta}_{adv}=\bm{z}_{adv}-\bm{z}$. This violates the definition of a Latent Consistency Model mentioned in ~\cref{consistency function defintion}. To address this disharmony, we introduce $\bm{z}^\ast_t$
\begin{equation}
    \bm{z}^\ast_{t}=\sqrt{\bar{\alpha}_{t}}\bm{z}+\sqrt{1-\bar{\alpha}_{t}}(\bm{\epsilon}+\bm{\delta}_{adv}),
\label{forward diffuse}
\end{equation}
by borrowing the idea of forward diffusion process, making $\bm{z}^\ast_{t}$ be an linear combination of $\bm{z}$ and $\bm{z}_{adv}$, worth to mention that these $\bm{\delta}_{adv}$ used in GAND are generated from latent space attack.
\begin{equation}
    \bm{\delta}_{adv}=\arg\max_\delta\mathcal{L}(C(\mathcal{D}(\mathcal{E}(\bm{x})+\bm{\delta})),\bm{y}).
\label{latent space attack}
\end{equation}
The $\bm{z}^\ast_t\rightarrow \bm{z}$ when $t\rightarrow 0$, and $\bm{z}^\ast_t\rightarrow \bm{\epsilon}+\bm{\delta}_{adv}$ when $t\rightarrow T$, which converges to the natural image when $t$ is small and matches with our goal, removing the adversarial noise and Gaussian noise concurrently from the shifted normal distribution. Hence, we eliminate disharmony between adversarial training and the original latent consistency distillation objective.

Then, as we do not give condition embeddings, we remove the $\omega$ in the formulation of estimation $\hat{\bm{z}}^{\Psi,\omega}_{t_{n}}$ to be:
\begin{equation}
     \hat{\bm{z}}^{\Psi}_{t_{n}}\leftarrow\bm{z}_{t_{n+k}}+\Psi(\bm{z}_{t_{n+k}},t_{n+k},t_n,\varnothing),
    \label{denoise step}
\end{equation}
where $\varnothing$ means null conditional embedding.

Furthermore, our goal is purifying images instead of image generation, we can train our LCM in a weaker constraint, where we only need the consistency function satisfying on $[0,t]$, $t<T$. Hence, we only simulate the $n$ of time step $t_n$ in $\mathcal{U}[1,(N-k)/2]$. This means that our LCM only satisfies the consistency function in the first half of the time steps.

In addition, inspired by ~\cite{kim2023consistency}, we add a Clear Image Guide (CIG) loss to further ensure our distillation process is training towards the proposed purification goal. CIG loss is given by:
\begin{equation}
    \mathcal{L}_{CIG}(\bm{\theta})=\mathbb{E}_{z,n}[d(f_\theta(\bm{z}^\ast_{t_{n+k}},\varnothing,t_{n+k}),\bm{z})].
    \label{cig}
\end{equation}
Therefore, our total distillation loss is the combination of GAND loss
\begin{equation}
\begin{split}
    \mathcal{L}_{GAND}(\bm{\theta},&\bm{\theta}^-)=\\ \mathbb{E}_{z,n}[&d(f_\theta(\bm{z}^\ast_{t_{n+k}},\varnothing,t_{n+k}),f_{\theta^-}(\hat{\bm{z}}_{t_n}^{\Psi},\varnothing,t_n))],
    \label{gand}
\end{split}
\end{equation}
and CIG loss as:
\begin{equation}
    \mathcal{L}_{Total}(\bm{\theta},\bm{\theta}^-)=\mathcal{L}_{GAND}(\bm{\theta},\bm{\theta}^-)+\lambda_{CIG}\mathcal{L}_{CIG}(\bm{\theta}),
    \label{total loss}
\end{equation}
$\lambda_{CIG}$ is used to ensure two losses are on the same scale.

The following theorem shows if the loss of GAND converges to an arbitrarily small number, the difference between the purified image and the clean image will be arbitrarily small. We can use this as guidance on how to choose our purification $t$ in the inference step since our method is reaching the theoretical optimal performance at a specific $t$.

\begin{figure}[t]
    \centering

    \includegraphics[width=\linewidth]{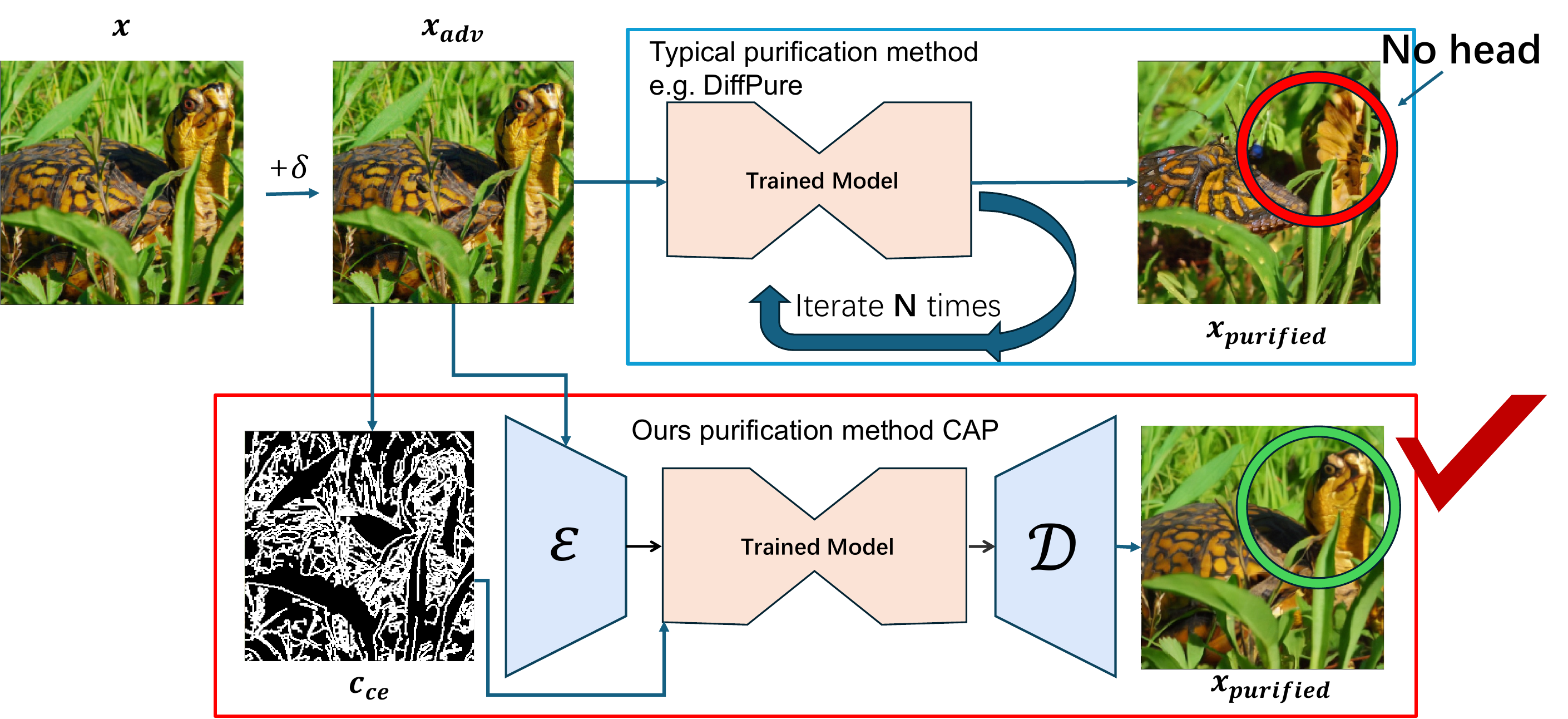}

    \caption{CAP used edge image of the adversarial image to control our purification process, maximizing the remaining semantic information of the purified image.}
    \label{fig:cap}

\end{figure}

\subsection{Controlled Adversarial Purification}
To address the first component we defined, we propose CAP (as shown in Fig.~\ref{fig:cap}) - a purification inference pipeline that utilizes visual prompts instead of text guidance. We chose visual prompts because text-based guidance can be vulnerable to caption semantic attacks~\cite{Xu_2019_CVPR}, which may compromise the accuracy of generated text prompts. Therefore, we opt for a more traditional and robust approach by using an unlearnable edge detection operator to generate edge guidance of the input adversarial sample.

\begin{table*}[tbp]
\centering

\caption{Accuracy (\%) results for ImageNet. The best models against different attack methods are in \textbf{bold}. The methods marked with * mean the data are borrowed from the original paper.}\label{tab:imagenet}

\input{floating/acc_imagenet}

\end{table*}

In purification process, we first encode the adversarial image $\bm{z}_{adv}=\mathcal{E}(\bm{x}_{adv})$ to latent space using pre-trained image encoder $\mathcal{E}$ and sample a random noise $\bm{\epsilon}\sim\mathcal{N}(\bm{0},\mathbf{I})$ in the dimension of the latent
space. Then, we diffuse the $\bm{z}_{adv}$ with predefined strength $t^\ast$, using forward latent diffusion process:
\begin{equation}
    \bm{z}_{adv}(t^\ast) = \sqrt{\Bar{\alpha}(t^\ast)}\bm{z}_{adv}+\sqrt{1-\Bar{\alpha}(t^\ast)}\bm{\epsilon}.
\end{equation}
Then, we purify $\bm{z}_{adv}(t^\ast)$ using our LCM trained by GAND (in Alg.~\ref{alg:adv_lcm}),  $f_\theta(\bm{z},\bm{c},t)$, $\bm{z}$ is image latent, $\bm{c}$ is the condition embedding (e.g., text, canny edge image) and $t$ is time step. The Latent Consistency Model has been introduced in ~\cref{latent consistency function}. The purified image latent comes from the latent consistency function, $\hat{\bm{z}}_{adv}^0= f_\theta(\bm{z}_{adv}(t^\ast),\bm{c}_{ce}, t^\ast)$, where $\bm{c}_{ce}$ means canny edge images which are provided by edge detection operators~\cite{canny1986computational}. Although ControlNet is an extra plug-in tool for Stable Diffusion, $\bm{c}_{ce}$ is not exactly an input of $f_\theta$, we treat it as condition embedding here for simplifying the equation. To further reduce the effect of the adversarial image, we remove the $c_{skip}(t)\bm{z}_{adv}(t)$ in LCM $f_\theta$ and denote this LCM as $f_\theta^-$, since this term will maintain most of the adversarial noise. Therefore, our purified image latent is actually:
\begin{equation}
\begin{split}
    \hat{\bm{z}}_{adv}^0&= f_\theta^-(\bm{z}_{adv}(t^\ast),\bm{c}_{ce}, t^\ast)\\&=c_\text{out}(t)\left(\frac{\bm{z}_{adv}-\sqrt{1-\Bar{\alpha}_t}\hat{\epsilon}_\theta(\bm{z}_{adv},\bm{c}_{ce},t)}{\sqrt{\Bar{\alpha}_t}}\right).
\end{split}
\label{f minus}
\end{equation}
Same as setting $c_{skip}(t)\equiv0$. Finally, We reconstruct the purified image by
the image decoder $\mathcal{D}$, $\hat{\bm{x}}_{adv}^0=\mathcal{D}(\hat{\bm{z}}_{adv}^0)$.

%% file: floating/acc_imagenet.tex
\begin{tabular}{l|c|cc|r}
\toprule
Defense                        & Attack Method       & Standard Accuracy       & Robust Accuracy          & Architecture \\ \midrule
Without defense                                   & untargeted PGD-100     & 80.55\%           & \ 0.01\%            & ResNet-50    \\
Without defense                                   & AutoAttack             & 80.55\%           & \ 0.00\%               & ResNet-50    \\
Without defense                                   & random-targeted PGD-40 & 82.33\%           & \ 0.04\%             & ResNet-152   \\ 
\midrule
\textbf{Diffusion Based Purification}&      &      & \\
\citet{wang2022guided}             & untargeted PGD-100     & 73.53\%           & 72.97\% & ResNet-50    \\
\citet{wang2022guided}             & random-targeted PGD-40 &  78.10\% & 77.86\% & ResNet-152   \\
DiffPure~\cite{nie2022diffusion}       & AutoAttack           & 75.77\%           & 73.02\%           & ResNet-50    \\ \midrule
\textbf{Adversarial Training}&      &      & \\
\citet{amini2024meansparse}* & AutoAttack          & \textbf{77.96\%}  & 59.64\% & ConvNeXt-L   \\
\citet{singh2024revisiting}*   & AutoAttack            & 77.00\% & 57.70\%           & ConvNeXt-L   \\ \midrule
\textbf{Hybrid}&      &      & \\
OSCP (ours)                            & untargeted PGD-100   &  \textbf{77.63\%}  & \textbf{73.89\%}  & ResNet-50    \\
OSCP (ours)                            & AutoAttack            & 77.63\%           & \textbf{74.19\%}  & ResNet-50    \\
OSCP (ours)                            & random-targeted PGD-40 & \textbf{79.81\%}  & \textbf{78.78\%}  & ResNet-152   \\ 
\bottomrule

\end{tabular}
\vspace{-2pt}

%% file: sec/5_exp.tex
\section{Experiments and Results}

\noindent \textbf{Training setting} Our adversarial latent consistency model is distilled from Stable Diffusion v1.5~\cite{rombach2022high}, which utilizes $\epsilon$-prediction~\cite{ho2020denoising}. We split the first 10,000 images and the remaining 40,000 images in the ImageNet~\cite{deng2009imagenet} valset as our testing set and training set, respectively. The GAND framework is trained on our training set, with all images resized to 512$\times$512 resolution. Training proceeds for 20,000 iterations with a batch size of 4, employing an initial learning rate of 8e-6 and a 500-step warm-up period. For the diffusion process, we employ DDIM-Solver~\cite{songdenoising} as the PF-ODE solver $\Psi$ with a skipping step $k = 20$ in Equation~\ref{denoise step}. Adversarial latents are generated using PGD-10 with $\epsilon = 0.03$ against a ResNet50 victim model, and the consistency improvement gradient weight $\lambda_{CIG}$ is set to 0.001. Notably, our experiments demonstrate that PEFT via LoRA is sufficient for achieving optimal performance.

\noindent \textbf{Attack Settings}
We conduct our evaluations on our testing set. Our evaluation encompasses diverse architectures, spanning traditional CNNs (ResNet 50 and 152~\cite{he2016deep}, WideResNet~\cite{zagoruyko2016wide}) and modern Vision Transformers (ViT-B~\cite{dosovitskiy2020image}, Swin-B~\cite{liu2021swin}).
For adversarial attacks, we employ standard $L_p$ norm-based methods including PGD~\cite{madry2018towards} and AutoAttack~\cite{croce2020reliable}. The PGD attacks are configured with $L_\infty$ bounds $\gamma \in \{4/255, 16/255\}$ and corresponding step sizes $\eta \in \{1/255, 0.025 \cdot 16/255\}$, while AutoAttack uses $L_\infty$ bound $\gamma = 4/255$. Here, PGD-$n$ denotes PGD attack with $n$ iterations.
In our evaluation metrics, we define standard accuracy as performance on clean data with our defense framework, robust accuracy as performance on adversarial data with our defense, and clean accuracy as performance on clean data without any defense. We fix the random seed of every experiment to avoid
the randomness.

\begin{table}[t]
\centering
\caption{Accuracy (\%) and attack success rate (ASR \%) results in various architectures, where the attack method is PGD-100, with budget $4/255$, attack step size 1/255.}\label{tab:transfer}
\input{floating/transferability}
\end{table}

\begin{table*}[ht]
\centering
\caption{Robust Accuracy (\%) on our method under Diff-PGD-10 attack $\gamma=8/255$ ($\eta=2/255$) on ImageNet~\cite{deng2009imagenet}.}
\label{tab:multistep}

\begin{tabular}{c|cccccc}
\toprule
Defence method & ResNet-50& ResNet-152&WideResNet-50-2&Vit-b-16& Swin-b& ConvNeXt-b\\
\midrule
DiffPure ($t^\ast = 100$)& 53.8\%      &   49.4\% & 52.2\%  &  16.6\%  & 45.1\% & 42.9\% \\
\midrule
Ours   ($t^\ast=250$)  &\textbf{59.0\%}     &    \textbf{56.5\%}& \textbf{57.9\%} & \textbf{34.1\%}  & \textbf{53.9\%} & \textbf{49.1\%} \\

\bottomrule
\end{tabular}
\vspace{-\belowdisplayskip}
\end{table*}

\subsection{Main Result}

Our method is regarded as a hybrid method, consisting of adversarial training and purification. Hence we compare our method with both categories of the defenses. Based on the provided attack setting, our method demonstrates significant computational efficiency, completing evaluations in 3 hours on an NVIDIA F40 GPU compared to GDMP's \cite{wang2022guided} 48 hours. As shown in Tab.~\ref{tab:imagenet}, diffusion based purification achieves satisfactory robust accuracy while sacrificing standard accuracy. On the other hand, adversarial training has little standard accuracy loss but the robust accuracy is unsatisfactory. Our method achieves promising results on both standard and robust accuracy with three different attack methods. We achieve robust accuracy improvements of 0.92\% and 1.17\% against PGD-100 ($\gamma=4/255$) and AutoAttack respectively, while improving from 77.86\% to 78.78\% against random targeted PGD-40 ($\gamma=16/255$). 
We empirically determine $t^\ast=200$ as the optimal purification strength.

\subsection{Analysis With Different Architectures}

While our GAND-trained LCM is initially optimized using adversarial examples generated for ResNet50, we evaluate its generalization capability across different architectures to address potential overfitting concerns. As shown in Tab.~\ref{tab:transfer}, our method maintains strong robust accuracy across architectures, with even the lowest performance of 71.6\% on ViT$_{b-16}$ demonstrating effective transfer. These experiments, along with subsequent ImageNet evaluations, are conducted on a representative subset of 500 images from our previously mentioned 10,000 images test set, as we observe consistent performance between these sample sizes.

\subsection{Defend Adaptive Attack}

In Tab.~\ref{tab:multistep}, we test our method under version 2 of Diff-PGD attack~\cite{xue2024diffusion}, which is a SOTA attack method against diffusion based purification. Our method boost the robust accuracy in Vit-b-16 from 16.6\% to 34.1\%. In ResNet-50, ResNet-152, WideResNet-50-2, the robust accuracy increase 5.2\%, 7.1\% and 5.7\% respectively. For robust accuracy in Swin-b and ConvNeXt, our method acheive 53.9\% and 49.1\%, which is 8.8\% and 6.2\% higher than DiffPure. Worth to mention that our method uses LCM model which is deterministic, and hence, Expectation over Transformation, which is usually use in other robust evaluation for DBP, is not needed here.

\subsection{Inference Time}

\begin{table}[tbp]
\centering
\caption{Inference time of purification models to purify one image on an NVIDIA F40 GPU. Notably, GDMP refers to \citet{wang2022guided}, DiffPure refers to \citet{nie2022diffusion}.}
\label{tab:inferencetime}
\resizebox{\linewidth}{!}{%
\input{floating/runtime}

}

\end{table}

Conventional diffusion-based purification methods are computationally intensive, making them impractical for real-time applications.
As shown in Tab.~\ref{tab:inferencetime}, our method achieves purification in just \textbf{0.1s} for ImageNet and \textbf{0.5s} for CelebA-HQ (1024 $\times$ 1024) on a NVIDIA F40 GPU, \textbf{independent of $t^\ast$} where the inference time of other method scales up depending on $t^\ast$. More detailed experiments can be found in the supplementary materials. Our method significantly speeds up the purification process, and this acceleration enables deployment in time-critical scenarios such as autonomous driving and purification in video.

\subsection{Face Recognition}

We further test our model on a subset of 1000 images from CelebA-HQ~\cite{liu2015faceattributes}, choosing purification step $t^\ast$ as 200, control scale 0.8, and utilizing the GAND weights trained on ImageNet. We defend against targeted PGD attacks that attempt to manipulate Arcface~\cite{deng2019arcface} (AF), FaceNet~\cite{schroff2015facenet} (FN), and MobileFaceNet~\cite{chen2018mobilefacenets} (MFN) to misidentify any input face as a specific target person. Our defense goal is to prevent this misidentification by ensuring the cosine similarity between the purified image embedding and the target image remains below the recognition threshold.
To validate the effectiveness of ImageNet-trained GAND weights on CelebA-HQ, we compare OSCP with CAP (OSCP without GAND) as shown in Tab.~\ref{tab:celeba}. The GAND weights significantly improve robust accuracy from 83.4\% to 86.8\% on Arcface and from 82.8\% to 84.9\% on MobileFaceNet. Notably, both our methods (CAP and OSCP) achieve an exceptional robust accuracy of 97.8\% on FaceNet.

\vspace{-1mm}

\begin{table}[t]
\centering
\caption{Robust accuracy (\%) for CelebA-HQ under targeted PGD-10 $L_\infty\gamma(\gamma=4/255)$, $\eta=$ 0.5*5/255. The best model is in \textbf{bold}.}
\label{tab:celeba}

\input{floating/celab}
\vspace{-1mm}
\end{table}

\subsection{Image Quality Assessment}
\begin{table}[tbp]
\centering
\caption{IQAs for $\epsilon=8/255$. $\uparrow$/$\downarrow$ indicate higher/lower value consists the better image quality. Notably, $\operatorname{IQA}(\cdot)$ indicates one of IQA we leverage, $\boldsymbol{x}$ refers to the clean image, $\boldsymbol{x}_{\text{adv}}$ refers to the adversarial sample, $\boldsymbol{x}_{\text{ours}}$ refers to the purified image by our method and $\boldsymbol{x}_{\text{DiffPure}}$ refers to purified image by DiffPure.}

\label{tab:iqa:eps4}
\begin{tabular}{llll}
\hline
                                                                      & LPIPS$\downarrow$            & PSNR$\uparrow$             & SSIM$\uparrow$             \\ \hline
$\operatorname{IQA}(\boldsymbol{x},\boldsymbol{x}_{\text{adv}})$      & 0.1547         & 34.21         & 0.8906         \\ \hline
$\operatorname{IQA}(\boldsymbol{x},\boldsymbol{x}_{\text{ours}})$     & \textbf{0.2370} & \textbf{24.13} & \textbf{0.7343} \\
$\operatorname{IQA}(\boldsymbol{x},\boldsymbol{x}_{\text{DiffPure}})$ & 0.2616         & 24.11         & 0.7155          \\ \hline
\end{tabular}

\end{table}

Regarding image quality assessments, we compare our method with DiffPure using a subset of 1000 ImageNet images from our testing set. We use PSNR to reflect the overall distortion level of images; SSIM~\cite{wang2004image} to indicate how human eyes perceive image structures; LPIPS~\cite{zhang2018unreasonable} to assesses perceptual similarity.
Both methods are configured with $t^\ast$=250 for fair comparison. As shown in Tab.~\ref{tab:iqa:eps4}, our method demonstrates superior performance across multiple metrics: LPIPS decreases from 0.2616 to 0.2370, while PSNR and SSIM improve from 24.11 to 24.13 and 0.7155 to 0.7343.
These improvements consistently indicate that our purified images maintain greater fidelity to the original inputs, validating the effectiveness of our purification approach. The visualization is shown in Fig. \ref{fig:iqa}.

\input{floating/IQA}

\begin{figure}[t]
    \centering
    \includegraphics[width=0.9\linewidth]{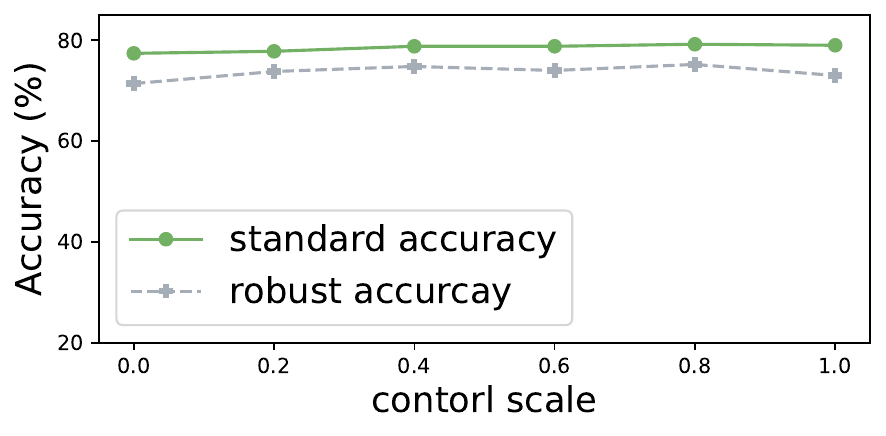}
    \caption{Performance of our method on different $t^\ast$ under PGD-100 $L_\infty \gamma$ $(\gamma = 4/255)$, $\eta=$ 0.01*4/255, where we evaluate on ResNet50 on ImageNet.}
    \label{fig:control_scale}
    \vspace{-1mm}
\end{figure}

\begin{table}[t]
    \centering
    \caption{Accuracy (\%) under PGD-100
$L_{\infty}\gamma$ ($\gamma$ = 4/255), $\eta=$ 0.01*4/255. The best result is in \textbf{bold}, the 2nd best is \underline{underlined}.}
    \label{tab:ab}
    \resizebox{0.65\linewidth}{!}{
    \begin{tabular}{c|c|c c} \toprule 
           CAP &  GAND &  Standard & Robust   \\ \midrule
         \ding{55} & \ding{55} & 75.0\% &72.4\%    \\ \midrule
         \ding{55} & \ding{51} & 74.0\% &72.6\%    \\  \midrule
         \ding{51} & \ding{55} & \textbf{77.0\%}&\underline{73.2\%}    \\  \midrule
        \ding{51}  & \ding{51} & \underline{76.8\%}&\textbf{75.0\%}    \\ \bottomrule
    \end{tabular}}
    \vspace{2mm}
\end{table}

\subsection{Ablation Studies}

We conduct comprehensive ablation studies on 500 images subset from our testing set. First, as shown in Fig.~\ref{fig:control_scale}, we optimize CAP's performance across different ControlNet conditioning scales, identifying 0.8 as the optimal value.
Tab.~\ref{tab:ab} demonstrates the individual and combined effects of our components. The baseline LCM-LoRA already shows promising performance, indicating its potential for adversarial purification. GAND-trained weights yield marginal improvements, while CAP integration increases standard and robust accuracy by 2\% and 0.8\% respectively. The full OSCP framework (combining both) significantly improves robust accuracy from 72.4\% to 75\%, validating our design choices. In Tab.~\ref{tab:cskip}, we analyze the impact of the $c_{skip}(t)\bm{z}_t$ term from~\cref{f minus}. While standard accuracy remains stable, removing this term improves robust accuracy, likely because $\bm{z}_t$ retains adversarial noise from the input image. This observation leads us to exclude this term from our LCM implementation.

Finally, Tab.~\ref{tab:training time} explores different training time step ranges: $\mathcal{U}[1,(N-k)/4]$, $\mathcal{U}[1,(N-k)/2]$, $\mathcal{U}[1,3(N-k)/4]$, and $\mathcal{U}[1,N-k]$ (original). Results show that smaller time step ranges maintain better performance, suggesting that fine-tuning LCM with smaller time steps is sufficient for purification tasks.

%% file: floating/transferability.tex
\begin{tabular}{l | l | c c c } 
\toprule
Architecture & Clean & ASR & Standard  & Robust \\
\midrule
WRN-50-2 &  82.6\%    & 100\%    &   77.6\%  &  75.2\% \\
\midrule

Vit-b-16 &  81.6\%     & 100\%    &  78.2\%  &   71.6\% \\
\midrule

Swin-b &    83.6\%      & 100\%    & 79.2\% & 77.8\% \\
\bottomrule
\end{tabular}
\vspace{1em}

%% file: floating/runtime.tex
\begin{tabular}{l|l|cc}
\toprule
Method          & Dataset  & $t^*\in$ [0,1000] & runtime (s) \\ \midrule
GDMP        & ImageNet & 250               & $\sim 9$           \\ \midrule
DiffPure     & ImageNet & 150               & $\sim 11$          \\ \midrule

OSCP(Ours)                              & ImageNet & 20/200/500/1000    & $\sim\textbf{0.1}$ \\
OSCP(Ours)                         & CelabA-HQ & 20/200/500/1000    & $\sim0.5$

\\ \bottomrule

\end{tabular}

%% file: floating/celab.tex
\begin{tabular}{l|cccc}
\toprule
Defense Method                       &  AF         &  FN   & MFN  \\ \midrule
Without defense                                   &  \ \ \ \ \ 0\%            &     \ \ 0.3\%    &    \ \ \ \ \ 0\%                 \\
GassianBlur ($\sigma$ = 7.0)            & \ \ 2.8\%            & 51.4\%         &\ \ 2.8\%          \\ 
\citet{das2018shield} (n = 60)        & 17.3\%           & 84.1\%         &27.6\%       \\   \midrule

CAP(Ours)                             & 83.4\%           & \textbf{97.8\%}         & 82.8\%   \\
OSCP(Ours)                           & \textbf{86.8\%} & \textbf{97.8\%}  & \textbf{84.9\%}
\\ \bottomrule
\end{tabular}

%% file: floating/IQA.tex
\begin{figure}[ht]
\centering
\small
\setlength{\tabcolsep}{1pt}
    \scalebox{0.95}{
\begin{tabular}[b]{cccc} 
    
        \includegraphics[width=0.12\textwidth]{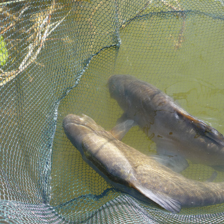}
     & 
    \includegraphics[width=0.12\textwidth]{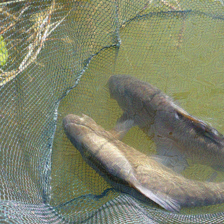}
     & 
    
        \includegraphics[width=0.12\textwidth]{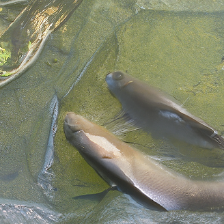}
        &
        \includegraphics[width=0.12\textwidth]{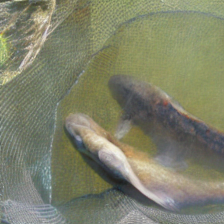}
    \\
    \includegraphics[width=0.12\textwidth]{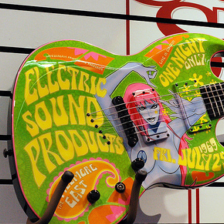}
     & 
    \includegraphics[width=0.12\textwidth]{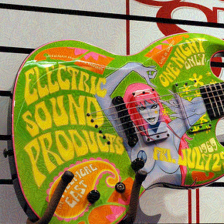}
     & 
    
        \includegraphics[width=0.12\textwidth]{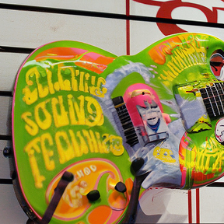}
        &
        \includegraphics[width=0.12\textwidth]{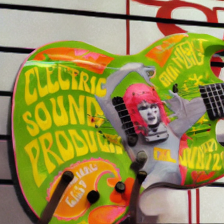}
     \\
    (a) & (b)  & (c)  & (d) 
\end{tabular}}
\caption{Visualization of the IQA experiment which compares with DiffPure and the proposed method. (a) Input image. (b) Adversarial image. (c) DiffPure. (d) Ours.
}
\label{fig:iqa}

\end{figure}

%% file: sec/6_conclu.tex
\vspace{-1mm}
\section{Conclusion and Discussion}

In this paper, we present One Step Control Purification (OSCP), a novel framework that addresses two critical challenges in diffusion-based adversarial purification: computational efficiency and semantic preservation. Our approach integrates the knowledge of pre-trained Diffusion Models with the proposed Gaussian Adversarial Noise Distillation (GAND) to achieve robust defense in a single inference step. Then as the inference framework, the Controlled Adversarial Purification (CAP), utilizing the unlearnable edge detection operator, effectively preserves semantic information of the input during purification.
Extensive experiments validate OSCP's effectiveness across diverse architectures and datasets. On ImageNet, we achieve 74.19\% robust accuracy against strong attacks while requiring only 0.1s per purification—a 100-fold speedup compared to existing methods. The framework demonstrates strong cross-domain generalization, from natural images to face recognition tasks, while our ablation studies confirm the synergistic benefits of combining GAND and CAP.

While our work represents significant progress in real-time adversarial defense, several promising directions remain, including the adaptive purification strength mechanisms and alternative control conditions. These advances could further enhance the applicability of diffusion-based defenses in time-critical scenarios.

\begin{table}[t]
\centering
\caption{Accuracy with and without $c_{skip}(t)\bm{z}_t$ term in LCM in our method under PGD-100 ($\gamma=8/255$), $\eta=2/255$ on ResNet50 on a subset of 500 images from our ImageNet testing set.}

\label{tab:cskip}
\resizebox{\linewidth}{!}{
\begin{tabular}{c|c|ccccc}
\toprule
$c_{skip}(t)\bm{z}_t$ & Acc. & $t^\ast$=20 & $t^\ast$=60 & $t^\ast$=100 & $t^\ast$=200 & $t^\ast$=250 \\
\midrule
\ding{55} & Robust & 43.4\% & \textbf{56.0\%}   & 59.4\% & \textbf{66.4\%} & \textbf{68.4\%} \\
\midrule
\ding{55} & Standard & 81.2\%	& 81.4\% & 81.2\% & 79.2\% & 76.6\% \\
\midrule

\ding{51} & Robust & \textbf{44.2\%} & 54.8\% & \textbf{60.4\%} & 66.0\%   & 67.8\% \\  
\midrule
\ding{51} & Standard & 81.2\%	& 81.4\% & 81.2\% & 79.2\% & 76.6\% \\
\bottomrule
\end{tabular}}
\end{table}

\begin{table}[t]
\centering
\small
\caption{Accuracy of our method under AutoAttack ($\gamma=8/255$) on ResNet50 on a subset of 1000 images from ImageNet on different sets of training time steps. ($t^\ast=200$)}
\label{tab:training time}
\begin{tabular}{c|lllll}
\toprule
 Acc. & $\frac{N-k}{4}$ & $\frac{N-k}{2}$ & $\frac{3(N-k)}{4}$ &  $N-k$ \\
\midrule
 Robust & 74.0\% & \textbf{74.3\%} & 73.7\% & 73.6\% \\
\midrule
 Standard & 78.8\% & \textbf{79.4\%} & 79.1\% & 79.1\% \\
\bottomrule
\end{tabular}

\end{table}

%% file: sec/X_suppl.tex
\clearpage

\maketitlesupplementary

\section{Appendix}

\subsection{Additional implementation details}

We implement our method with Pytorch~\cite{paszke2019pytorch} and Diffusers. we fix the random seed of PyTorch’s generator as 100 for reproducibility~\cite{picard2021torch}.

For all implementations, we use the standard version of AutoAttack, which is the same in both the main paper and the appendix.

We leverage \href{https://huggingface.co/latent-consistency/lcm-lora-sdv1-5}{LCM-LoRA}~\cite{luo2023lcm} and \href{https://huggingface.co/h1t/TCD-SD15-LoRA}{TCD-LoRA}~\cite{zheng2024trajectory} from their HuggingFace repositories.

In terms of training with our proposed GAND, we use the \href{https://huggingface.co/runwayml/stable-diffusion-v1-5}{Stable Diffusion v1.5} (SD15) as the teacher model.

We borrow a part of code and pretrained weights from \href{https://github.com/CGCL-codes/AMT-GAN}{AMT-GAN} when we do experiments on CelebA-HQ. We also conduct a series of IQAs to evaluate the quality of purified images. To be specific, we leverage:
\begin{itemize}
\item PSNR: Range set to 1, aligning with PyTorch's image transformation.
\item \href{https://github.com/Po-Hsun-Su/pytorch-ssim}{SSIM}~\cite{wang2004image}: Gaussian kernel size set to 11.
\item \href{https://github.com/richzhang/PerceptualSimilarity}{LPIPS}~\cite{zhang2018unreasonable}: Utilizing VGG \citep{simonyan2014very} as the surrogate model.
\end{itemize}

Experimenting our method in defending 
Fog~\cite{kaufmann2019testing}: We set the number of iterations to 10, $\epsilon$ to 128,
and step size to 0.002236.
Snow~\cite{kaufmann2019testing}: We set the number of iterations to 10, $\epsilon$ to
0.0625, and step size to 0.002236.
$L_2$-PGD: We set the number of iterations to 100, $\epsilon$ to 0.5, and step size to 0.1.

It is worth noting that we borrow the implementation of PSNR from \href{https://pytorch.org/torcheval/stable/}{TorchEval}.
Unless mentioned, all reproducibility-related things follow the above.

\subsection{Proofs}

We are going to provide some simple proofs for things we have claimed, including
\begin{enumerate}
    \item $\bm{z}^*_t\rightarrow \bm{z}$ when $t\rightarrow 0$, 
    \item $\bm{z}^*_t\rightarrow \bm{\epsilon}+\bm{\delta}_{adv}$ when $t\rightarrow T$
    \item $f_\theta(\bm{z}_{adv}(t),\varnothing,t)\rightarrow \bm{z}_{adv}$ when $t\rightarrow 0$
    \item $f_\theta(\bm{z}_{adv}(t),\varnothing,t)-f_\theta(\bm{z}(t),\varnothing,t)\rightarrow\bm{\delta}_{adv}$ when $t\rightarrow 0$
\end{enumerate}
Lemma. If $X\sim\mathcal{N}(\mu,\sigma^2)$ and $\sigma^2\rightarrow0$, then $X\rightarrow\mu$

Proof. For any $\epsilon>0$,
\begin{equation}
\begin{split}
     P(\|X-\mu\|\geq\epsilon)
     &= P(\|Z\|\geq\epsilon)  \text{\ \ \ \ $Z\sim\mathcal{N}(0,\sigma^2)$}\\
     &\leq \frac{E(X^2)}{\epsilon^2} \\
     &=\frac{Var(X)+(E(X))^2}{\epsilon^2}\\
     &=\frac{\sigma^2}{\epsilon^2}\\
     &\rightarrow0.
\end{split}
\end{equation}
The first line to the second line is true by Markov's inequality. Hence, we prove that normal distribution with a vanishing variance will converge in probability to its mean.

Proof 1. $\beta_t$ is increasing sequence in $t\in\{0,1,\cdots,T-1, T\}$ in range $(0,1)$, then we have $\alpha_t=1-\beta_t$ is decreasing sequence in $t\in\{0,1,\cdots,T-1, T\}$ in range $(0,1)$, $\bar{\alpha}_t$ is decreasing step function in range $(0,1)$, further assume $\beta_0$ is a arbitrarily small number:
\begin{equation}
    \begin{split}
        \lim_{t\rightarrow 0}\bm{z}_t^\ast&=\lim_{t\rightarrow 0}\sqrt{\bar{\alpha}_t} \bm{z}+\sqrt{1-\bar{\alpha}_t}(\bm{\epsilon}+\bm{\delta}_{adv})
        \\
        &=\sqrt{\alpha_0}\bm{z}+\sqrt{1-\alpha_0}(\bm{\epsilon}+\bm{\delta}_{adv})
        \\
        &=\sqrt{1-\beta_0}\bm{z}+\sqrt{\beta_0}\bm{\epsilon}+\sqrt{\beta_0}\bm{\delta}_{adv}\\
        &\rightarrow\bm{z}+\sqrt{\beta_0}\bm{\epsilon}.
    \end{split}
\end{equation}
The third line to the fourth line is true by assumption on $\beta_0$, and since we know $\bm{z}+\sqrt{\beta_0}\bm{\epsilon}\sim\mathcal{N}(\bm{z},\beta_0\bm{I})$ and $\beta_0$ is vanishing. By lemma, we have $\bm{z}^\ast_t\rightarrow\bm{z}$.

Proof 2. $\beta_t$ is increasing sequence in $t \in \{0,1,\cdots,T-1, T\}$ in range $(0, 1)$, then we have $\alpha_t$ is decreasing sequence in
$t \in \{0,1,\cdots,T-1, T\}$ in range $(0, 1)$, $\bar{\alpha}_t$ is decreasing step function in range $(0, 1)$,  further assume $T$ is a arbitrarily large number and $\beta_{T/2}>c$ where constant $c\in(0,1)$, consider $\bm{\delta}_{adv}$ as constant.

\begin{equation}
    \begin{split}
        \bar{\alpha}_T&=\prod_{t=0}^{T}\alpha_t \\
        &=\prod_{t=0}^{T}(1-\beta_t)\\
        &=\prod_{t=0}^{T/2}(1-\beta_t)\prod_{t'=T/2+1}^{T}(1-\beta_{t'})\\
        &\leq(1-c)^{T/2}\prod_{t=0}^{T/2}(1-\beta_t)\\
        &\rightarrow 0,\\
        \lim_{t\rightarrow T}\bm{z}_t^\ast&=\lim_{t\rightarrow T}\sqrt{\bar{\alpha}_t} \bm{z}+\sqrt{1-\bar{\alpha}_t}(\bm{\epsilon}+\bm{\delta}_{adv})\\
        &=\bm{\epsilon}+\bm{\delta}_{adv}\sim\mathcal{N}(\bm{\delta}_{adv},\bm{I}).
    \end{split}
\end{equation}
$(1-\beta_0)^{T+1}$ converge to 0 by assuming T is arbitrarily large.

Proof 3. $\beta_t$ is increasing sequence in $t\in\{0,1,\cdots,T-1, T\}$ in range $(0,1)$, then we have $\alpha_t$ is linear decreasing sequence in $t\in\{0,1,\cdots,T-1, T\}$ in range $(0,1)$, $\bar{\alpha}_t$ is decreasing step function in range $(0,1)$, further assume $\hat{\epsilon}$ has standard Gaussian distribution and $\beta_0$ is a arbitrarily small number,

\begin{equation}
\begin{split}
     &\lim_{t\rightarrow 0}f_\theta(\bm{z}_{adv}(t),\varnothing,t)\\
     =&\lim_{t\rightarrow 0}\frac{\sigma^2}{t^2+\sigma^2}\bm{z}_{adv}(t)\\
     &+\frac{t^2}{\sqrt{t^2+\sigma^2}}\left(\frac{\bm{z}_{adv}(t)-\sqrt{1-\Bar{\alpha}_t}\hat{\bm\epsilon}_\theta(\bm{z}_{adv}(t),\bm{c},t)}{\sqrt{\Bar{\alpha}_t}}\right)\\
    =&\lim_{t\rightarrow 0}\frac{\sigma^2}{t^2+\sigma^2}\bm{z}_{adv}(t)\\
    &+\lim_{t\rightarrow 0}\frac{t^2}{\sqrt{t^2+\sigma^2}}\left(\frac{\bm{z}_{adv}(t)-\sqrt{1-\Bar{\alpha}_t}\hat{\bm\epsilon}}{\sqrt{\Bar{\alpha}_t}}\right)\\
    =&\lim_{t\rightarrow 0}\sqrt{\bar{\alpha}_t}\bm{z}_{adv}+\sqrt{1-\bar{\alpha}_t}\bm{\epsilon}-\lim_{t\rightarrow 0}\frac{t^2}{\sqrt{t^2+\sigma^2}}\left(\frac{\sqrt{1-\Bar{\alpha}_t}\hat{\bm\epsilon}}{\sqrt{\Bar{\alpha}_t}}\right)\\
    =&\sqrt{\alpha_0}\bm{z}_{adv}+\sqrt{1-\alpha_0}\bm{\epsilon}-\lim_{t\rightarrow 0}\frac{t^2}{\sqrt{t^2+\sigma^2}}\left(\frac{\sqrt{1-\alpha_0}\hat{\bm\epsilon}}{\sqrt{\alpha_0}}\right)\\
    =&\bm{z}_{adv}+\sqrt{\beta_0}\bm{\epsilon}-\lim_{t\rightarrow 0}\frac{t^2}{\sqrt{t^2+\sigma^2}}\left(\sqrt{\beta_0}\right)\hat{\bm\epsilon}\\
    =&\bm{z}_{adv}+\lim_{t\rightarrow 0}\sqrt{\left(1+\frac{t^4}{t^2+\sigma^2}\right)\beta_0}\bm{\epsilon}.
\end{split}
\end{equation}
The second last line is from assumption on $\beta_0$ and the last line is from the property of normal distribution and assumption on $\hat{\bm{\epsilon}}$. Finally, by the fact that $\lim_{t\rightarrow 0}\sqrt{\left(1+\frac{t^4}{t^2+\sigma^2}\right)\beta_0}=0$ and lemma we have proved, we prove $f_\theta(\bm{z}_{adv}(t),\varnothing,t)\rightarrow\bm{z}_{adv}$.

 Proof 4. Following a similar way in proof 3, we have 
\begin{equation}
\begin{split}
     &\lim_{t\rightarrow 0}f_\theta(\bm{z}(t),\varnothing,t)\\
     =&\bm{z}+\lim_{t\rightarrow 0}\sqrt{\left(1+\frac{t^4}{t^2+\sigma^2}\right)\beta_0}\bm{\epsilon},
     \\
     &\lim_{t\rightarrow 0}f_\theta(\bm{z}_{adv}(t),\varnothing,t)-f_\theta(\bm{z}(t),\varnothing,t)\\
     =&\bm{z}_{adv}-\bm{z}+\lim_{t\rightarrow 0}\sqrt{2\left(1+\frac{t^4}{t^2+\sigma^2}\right)\beta_0}\bm{\epsilon}\\
     \rightarrow&\bm{z}_{adv}-\bm{z}=\bm{\delta}_{adv}.\\
\end{split}
\end{equation}

The last line uses the definition of $\delta_{adv}$.

\subsection{Experiment}

\begin{figure}[htbp]

    \includegraphics[width=0.9\linewidth]{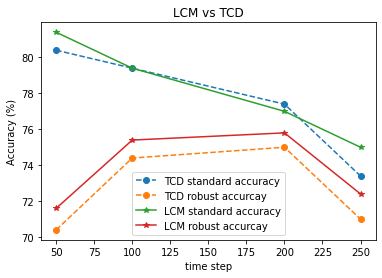}
    
    \caption{Accuracy (\%) using LCM and TCD model as defense model under $L_\infty$-PGD attack on ImageNet.}
    \label{fig:tcd}
\end{figure}

In~\cref{fig:tcd}, we test the robustness of another model that can do few steps generation, Trajectory consistency distillation (TCD)~\cite{zheng2024trajectory}, which can also generate an image in one step. We can see that using LCM as a purification model is generally more robust than using TCD. Also, the standard accuracy of the two models is similar. Therefore, we choose LCM as our backbone model for purification instead of TCD.

\begin{figure}[htbp]

    \includegraphics[width=0.9\linewidth]{figs/supp/time_step_effect_accuracy.pdf}
    
    \caption{Accuracy (\%) on different purification time step $t^*$ on our method. Two figures have the same attack setting, PGD-100 $L_\infty \gamma$ $(\gamma = 4/255)$, step size 0.01 * 4/255, where both we evaluate on ResNet50 on 500 images subset of ImageNet.}
    \label{fig:optimal step}
\end{figure}

\begin{figure*}[ht]
    \centering
    \includegraphics[width=\linewidth]{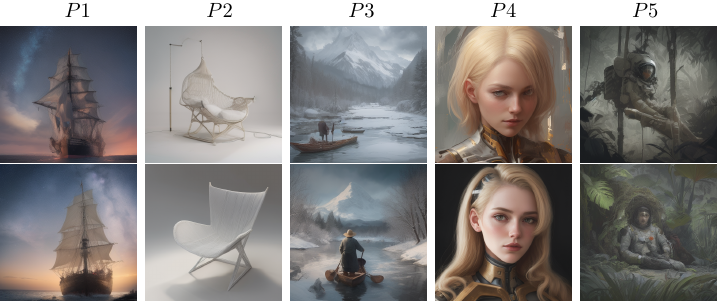}
\caption{Visualization of T2I (text to image) generation in different text prompts making use of our GAND weight. P1: A gorgeous ship sails under a beautiful starry sky. P2: Cradle chair, a huge feather, sandy beach background, minimalism, product design, white background, studio light, 3d, iso 100, 8k. P3: In a serene, snowy landscape, an elderly man in a straw raincoat and hat fishes alone on a small wooden boat in a calm, cold river. Surrounded by snow-covered trees and mountains, the tranquil scene conveys harmony with nature and quiet isolation. Masterpiece. Chinese art. extreme details. P4: Self-portrait oil painting, a beautiful cyborg with golden hair, 8k. P5: Astronauts in a jungle, cold color palette, muted colors, detailed, 8k}
    \label{fig:t2i}
\end{figure*}

In ~\cref{fig:optimal step}, we show the experiment for choosing the purification time step on ImageNet. Our model achieved the best performance at $t^\ast=200$ in both standard accuracy and robust accuracy. Hence, we decide to choose $t^\ast=200$ for our method on ImageNet.

\begin{table}[tb]
\centering
\caption{Inference time (s) of our method on different inference time and resolution}

\resizebox{\linewidth}{!}{
\begin{tabular}{c|c|lllll}
\toprule
Method & Resolution & $t^\ast=20$ & $t^\ast=60$ & $t^\ast=100$ & $t^\ast=250$ & $t^\ast=500$ \\
\midrule
DiffPure & 256$\times$256 & $\sim2$ & $\sim6$   & $\sim10$ & $\sim24$ & $\sim45$ \\

DiffPure & 512$\times$512  & $\sim6$	& $\sim17$ & $\sim27$ & $\sim70$ & $\sim140$ \\

DiffPure & 1024$\times$1024 & $\sim30$ & $\sim85$ & $\sim145$ & $\sim360$   & $\sim720$ \\
\midrule

Ours & 256$\times$256 & $\sim0.05$ & $\sim0.05$   & $\sim0.05$ & $\sim0.05$ & $\sim0.05$ \\

Ours & 512$\times$512  & $\sim0.1$	& $\sim0.1$ & $\sim0.1$ & $\sim0.1$ & $\sim0.1$ \\

Ours & 1024$\times$1024 & $\sim0.5$ & $\sim0.5$ & $\sim0.5$ & $\sim0.5$   & $\sim0.5$ \\  

\bottomrule
\end{tabular}}
\label{tab:appendex_time}
\end{table}

In~\cref{tab:appendex_time}, we further test the inference time of our method on three resolutions without resizing on an NVIDIA F40 GPU. This result showcases that if our GAND weights are trained on those resolutions, what will the inference time of our method be. As shown in Tab.~\ref{tab:appendex_time}, DiffPure takes around 12 minutes to purify a 1024$\times$1024 image when $t^\ast$ is chosen to be 500. Meanwhile, our method only takes 0.5 seconds to purify a 1024$\times$1024 image on any $t^\ast$, showing the potential of our method for purification in high-resolution images since our method does the purification in the latent space, which is much more efficient than pixel space.

\begin{figure}[tb]

    \centering
    \includegraphics[width=\linewidth]{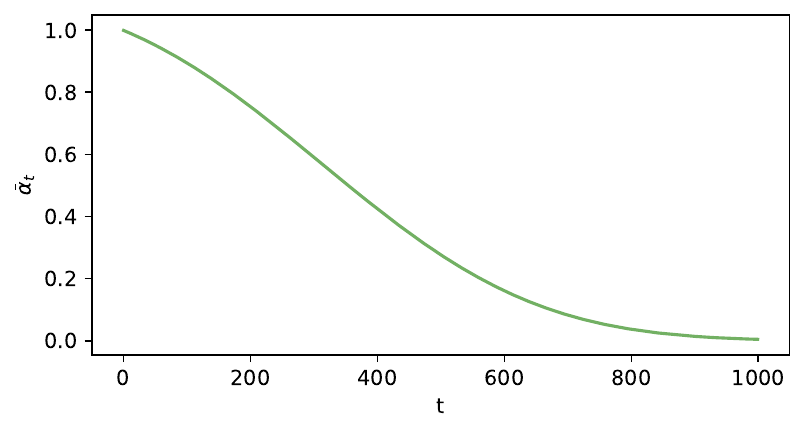}
    \caption{Visualization of how the term $\Bar{\alpha}_t$ change by t}
    \label{fig:beta_scheduler}
\end{figure}

In~\cref{fig:beta_scheduler}, we show how $\Bar{\alpha}_t$ variate by t in the \href{https://huggingface.co/docs/diffusers/api/schedulers/lcm}{LCM scheduler}. We can see that the value of $\bar{\alpha}_t$ starts at almost 1 ($t=0$) and decreases to almost 0 ($t=1000$), which meets with the assumptions we have made in our proofs.

\begin{figure*}
\centering
\begin{tabular}[b]{cccc}
    \includegraphics[width=0.24\textwidth]{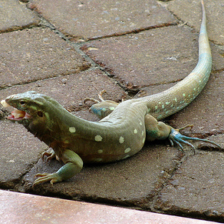}
     &
    \vspace{-1em}
    \hspace{-1em}
    \includegraphics[width=0.24\textwidth]{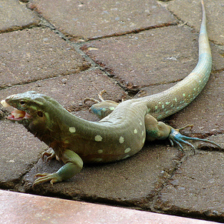}
     & 
     \hspace{-1em}
    
        \includegraphics[width=0.24\textwidth]{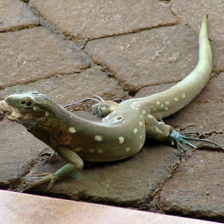}
        &
        \hspace{-1em}
        \includegraphics[width=0.24\textwidth]{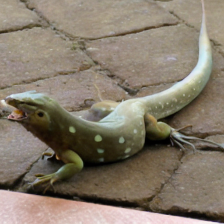}
     \\
         \includegraphics[width=0.24\textwidth]{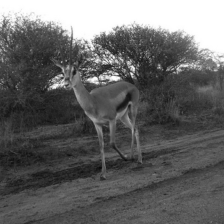}
     & 
     \hspace{-1em}
    \includegraphics[width=0.24\textwidth]{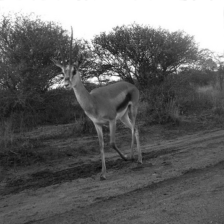}
     & 
     \hspace{-1em}

        \includegraphics[width=0.24\textwidth]{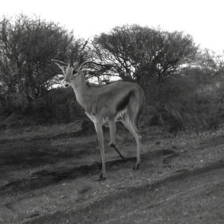}
        &
        \hspace{-1em}
        \includegraphics[width=0.24\textwidth]{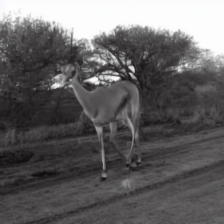}
     \\    \includegraphics[width=0.24\textwidth]{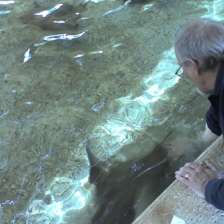}
     &
     \hspace{-1em}
    \includegraphics[width=0.24\textwidth]{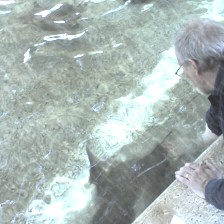}
     & 
     \hspace{-1em}
    
        \includegraphics[width=0.24\textwidth]{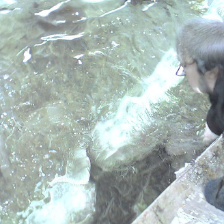}
        &
        \hspace{-1em}
        \includegraphics[width=0.24\textwidth]{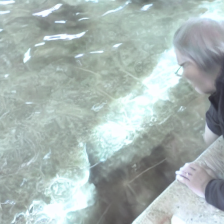}
     \\
     \includegraphics[width=0.24\textwidth]{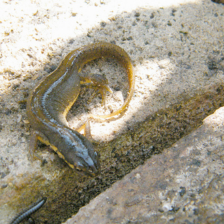}
     & 
     \hspace{-1em}
    \includegraphics[width=0.24\textwidth]{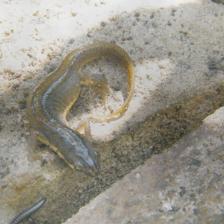}
     & 
     \hspace{-1em}
    
        \includegraphics[width=0.24\textwidth]{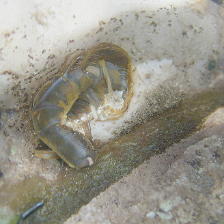}
        &
        \hspace{-1em}
        \includegraphics[width=0.24\textwidth]{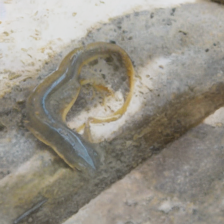}
    \\
    \midrule
    \includegraphics[width=0.24\textwidth]{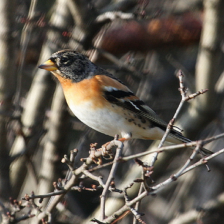}
     & 
     \hspace{-1em}
    \includegraphics[width=0.24\textwidth]{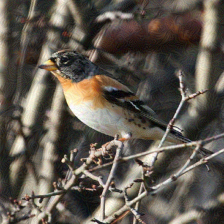}
     & 
     \hspace{-1em}
    
        \includegraphics[width=0.24\textwidth]{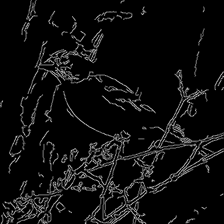}
        &
        \hspace{-1em}
        \includegraphics[width=0.24\textwidth]{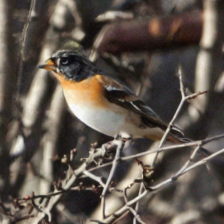}
    \\
    (a) & (b)  & (c)  & (d) 

\end{tabular}
\caption{Visualization of the experiment $L_2$-PGD, StAdv, Snow, Fog respectively, which compares with DiffPure and the proposed method. (a) Input image. (b) Adversarial image. (c) DiffPure. (d) Ours. The last row presents the proposed method (a) Input image (b) Adversarial image (c) Edge image (d) Purified Image
}
\label{fig:unseen}
\end{figure*}

In~\cref{tab:defense unseen}, we conducted four more attack methods on 500 images subset of ImageNet; we can see that our method is generally more robust than DiffPure, which is 1\% and 2.6\% higher than DiffPure when defending $L_2$-PGD and Snow. A slightly higher robust accuracy in defending StAdv and the same robust accuracy in defending Fog. Also, our method has a higher standard of accuracy, which is 1.6\% higher than DiffPure, showing that our method prevents semantic loss in the purification process. In~\cref{fig:unseen}, we provide more visual examples.

Although our goal is purification, it is still interesting for us to visualize the image generated by our GAND weight. We use LCM-LoRA and change their LoRA to our GAND LoRA weight. Then, we generate pictures by text prompt P1 (ship), P2 (chair), P3 (elderly man), P4 (Girl), and P5 (Astronauts); the full-text prompt is shown in the caption of ~\cref{fig:t2i}. We can see that our LCM-LoRA (GAND) still maintains generability; this is a surprising result and shows that our objective might be useful for some img2img tasks if people edit our objective correctly for their task.

\begin{table}[tb]
\centering
\caption{Standard accuracy and robust accuracies against unseen threat models on ResNet-50, $L_2$-PGD, StAdv, Snow, Fog}
\resizebox{\linewidth}{!}{
\begin{tabular}{c|lllll}
\toprule
Method & Standard & $L_2$-PGD & StAdv & Snow & Fog \\
\midrule
DiffPure &80.8\% &     80.6\% &69.4\%   & 77.2\%  & \textbf{77.8}\% \\
Ours &    \textbf{82.4}\% &      \textbf{81.6}\%       &\textbf{69.8}\% & \textbf{79.8}\%  & \textbf{77.8}\% \\
\bottomrule
\end{tabular}}
\label{tab:defense unseen}
\end{table}

